%% file: iclr2020_conference.tex
\documentclass{article} % For LaTeX2e
\usepackage{iclr2020_conference,times}

\input{math_commands.tex}

\usepackage{hyperref}
\usepackage{url}

\usepackage{wrapfig}
\usepackage{authblk}

\title{A Data and Compute Efficient Design \\ For Limited-Resources Deep Learning}

\author[2,1]{\textbf{Mirgahney Mohamed
	\thanks{Equal contribution}\ \ 
	\thanks{Work done during internship at Qualcomm AI Research}\ \ 
}}
\author[1]{\textbf{Gabriele Cesa\footnote[1]{} \ }}
\author[1]{\textbf{Taco S. Cohen}}
\author[1]{\textbf{Max Welling}}
\affil[1]{Qualcomm AI Research\thanks{Qualcom AI Research is an initiative of Qualcomm Technologies, Inc.}\protect\\Qualcomm Technologies Netherlands B.V.\protect\\\small{\texttt{\{gcesa,tacos,mwelling\}@qti.qualcomm.com}}}
\affil[2]{African Institute for Mathematical Sciences\protect\\\small{\texttt{mmerghaney@aimsammi.org}}}

\iclrfinalcopy % Uncomment for camera-ready version, but NOT for submission.

\begin{document}

\maketitle

\input{sections/00-abstract}
\input{sections/01-introduction}

\input{sections/02-relatedworks}
\input{sections/03-equivariance}

\input{sections/04-method}

\input{sections/05-experiments}

\input{sections/06-conclusions}

\subsubsection*{Acknowledgments}
We would like to thank Markus Nagel and Andrii Skliar for helping with quantization.
We are also thankful to Davide Belli, Andrii Skliar and Tijmen Blankevoort for proof reading this manuscript and the helpful feedback.

\bibliography{iclr2020_conference}
\bibliographystyle{iclr2020_conference}

\end{document}

%% file: math_commands.tex
\usepackage{amsmath,amsfonts,bm}

\usepackage{gensymb}
\usepackage{bm}

\DeclareMathOperator*{\R}{\mathbb{R}}

\newcommand{\bvec}{{\bm b}}

\newcommand{\xvec}{{\bm x}}
\newcommand{\yvec}{{\bm y}}

\usepackage{centernot}

\usepackage{mathtools}

\DeclareMathSymbol{\shortminus}{\mathbin}{AMSa}{"39}

%% file: sections/00-abstract.tex
%!TEX root=../iclr2020_conference.tex

\begin{abstract}

Thanks to their improved data efficiency, equivariant neural networks have gained increased interest in the deep learning community.
They have been successfully applied in the medical domain where symmetries in the data can be effectively exploited to build more accurate and robust models.
To be able to reach a much larger body of patients, mobile, on-device implementations of deep learning solutions have been developed for medical applications.
However, equivariant models are commonly implemented using large and computationally expensive architectures, not suitable to run on mobile devices. 
In this work, we design and test an equivariant version of MobileNetV2 and further optimize it with model quantization to enable more efficient inference.
We achieve close-to state of the art performance on the Patch Camelyon (PCam) medical dataset while being more computationally efficient. 

\end{abstract}

%% file: sections/01-introduction.tex
%!TEX root=../iclr2020_conference.tex

\section{Introduction}
\label{sec:intro}

Deep learning has recently moved closer to edge devices~\cite{8763885}, creating opportunities for many new applications where data needs to be analyzed in real time.
However, this development opens new challenges to improve power consumption, computational efficiency and memory footprint ~\cite{Rallapalli2016AreVD}. 
Many efforts have been directed toward compute, memory, and power efficient architecture design~\cite{Sandler2018, cai2018proxylessnas, Tan2019, Howard2019}.
Additionally, methods like compression~\cite{NIPS2015_5784, han2015compression, kuzmin2019taxonomy} and quantization~\cite{MellerFAG19samesame, Nagel2019, cai2020zeroq} have become popular.

These works become especially important in developing countries where the constrained resources make deploying state of the art models challenging~\cite{ml4d, sinha2019quantized}. 
Computer vision and deep learning algorithms can provide low-cost solutions where human experts are not available~\cite{Wahle000798}.
For instance, they can power automatic systems which help doctors in performing diagnosis or can be combined with drones to perform aerial imaging, e.g. to monitor disaster areas~\cite{Kyrkou_2019_CVPR_Workshops}.
\cite{pmlr-v56-Quinn16} develop a system for point-of-care diagnostics which uses mobile phones and microscopes to automate the diagnosis of different diseases.

In many cases, computational power is not the only limiting resource: gathering large quantities of labelled data is often prohibitively expensive. 
In this context, equivariance has been found to be a useful design choice, improving data efficiency through built-in knowledge about the symmetries of the problem \cite{Cohen2016,Worrall2017-HNET,Weiler2018-STEERABLE}.
Equivariant networks guarantee pre-determined transformations of their outputs under corresponding transformations of the input signals, enabling them to easily generalize over transformed signals. 
For these reasons, they have been successfully applied to medical imaging as well as aerial imaging, where symmetries are common ~\cite{bekkers2018roto,Veeling2018-qh,winkels3DGCNNsPulmonary2018,Li2018, Chidester2019,Dieleman2016-CYC,Hoogeboom2018-HEX}. 

Unfortunately, the demands of data and compute efficiency can be at odds.
Indeed, equivariant networks often rely on expensive architectures.
Their data efficiency is also often exploited to build larger models without overfitting. 
However, this is not always affordable in real world applications, especially under computational constraints as on handheld devices.
Moreover, for very small model sizes, increased weight sharing implied by equivariance can limit the number of distinct learnable filters too much, potentially harming the expressiveness of the model.
It is therefore not clear yet if equivariance can still be combined with efficient architecture design.
In this work:
(i) We show that equivariance is a useful design choice even with limited computational resources and in small model regime.
(ii) We show that quantization can be used to increase efficiency without harming equivariance, preserving the stability of the model.
To the best of our knowledge, this work is the first to combine quantization with equivariant~networks.

%% file: sections/02-relatedworks.tex
%!TEX root=../iclr2020_conference.tex

\section{Related Works}
\label{sec:relatedworks}

Equivariance offers a principled way to design models when the problem of interest presents certain symmetries.
In particular, it has proven very successful in image processing and, therefore, it has been extensively studied in this context.
\cite{Cohen2016} generalize convolutional networks beyond translations to arbitrary discrete groups but only consider ${\pi\over2}$ rotations.
Following works \cite{Weiler2018-STEERABLE, bekkers2018roto, cheng2018rotdcf,bekkers2020bspline} extend group convolutions to arbitrary discrete rotations.
\cite{Worrall2017-HNET} achieve continuous rotation equivariance using steerable filters.
\cite{Cohen2017-STEER} describe steerable networks for finite groups, covering group convolution as a special case.
\cite{Weiler2019} extend steerable networks to all planar isometries.
See \cite{Kondor2018-GENERAL,generaltheory} for further generalizations.

Equivariance has recently been applied in a number of medical imaging problems.
\cite{Veeling2018-qh} builds a reflection and ${\pi\over2}$ rotation equivariant DenseNet~\cite{Huang2017} for histopatological tissues classification.
The fully-convolutional design allows them to also use this model for efficient segmentation of whole-slide images.
Following works exploit equivariance to additional rotations~\cite{bekkers2020bspline} or scale~\cite{worrallDeepScale} on the same task.
In the context of segmentation, most works~\cite{Linmans2018, Li2018, Chidester2019} implement versions of U-Net~\cite{Ronneberger2015} only equivariant to ${\pi\over2}$ rotations and reflections.

The interest in bringing deep learning on edge devices has inspired new research to improve the efficiency of the models.
Recent works design new models optimizing both performances and number of FLOPS required~\cite{Sandler2018, cai2018proxylessnas, Tan2019, Howard2019}. 
Other lines of research, instead, focus on optimizing existing models to reduce their cost.
In particular, quantization involves reducing the precision used to store the weights and the activations, typically to $8$-bit integers~\cite{Jacob2017}.
To preserve the full-precision performances, it is often necessary to change the architecture~\cite{Sheng2018}, perform additional training~\cite{Ullrich2017,Louizos2018} or adapt the training procedure~\cite{ZhoWu16Dorefa}.
\cite{Krishnamoorthi2018,Nagel2019} propose solutions which do not require data or additional training.
In this work, we adopt the data-free quantization methods from~\cite{Nagel2019}. 

%% file: sections/03-equivariance.tex
%!TEX root=../iclr2020_conference.tex

\section{Background on Equivariant Networks}
\label{sec:equivariance}

We interpret a neural network $\Phi$ as a sequence of layers $\{\phi_i\}_{i=1}^n$, each mapping from an input feature space $F_{i-1}$ to an output feature space $F_i$.
A feature space is the vector space containing the features produced by a layer of the network.
In many cases, we have prior knowledge about the symmetries of the problem. 
For instance, we know that medical images may appear with arbitrary orientation and that their labels are independent from their rotations.
These symmetries are modeled by a group $G$ and its action 
on the data, e.g. a rotation of the images.
It is therefore desirable to build a neural network $\Phi$ which guarantees the following property (\textbf{equivariance}):
\begin{equation}
\forall g \in G, f \in F_0 \quad \Phi(g\cdot f) = g \cdot \Phi(f)
\label{eq:equivariance_property}
\end{equation}
Most approaches achieve this by enforcing equivariance at each layer $\phi_i$ of the network.
This requires the definition of an action of the group on each intermediate feature space $F_i$ during the design of the model.
Each layer, then, needs to satisfy the equivariance property Eq.~\eqref{eq:equivariance_property} with respect to the group action on its own input and output feature spaces.
In the case of $2D$-convolutional neural networks, a feature map $f \in \R^{c\times h \times w}$ can be interpreted as a \emph{feature field}, i.e. a function
\footnote{The input images and features are assumed to be continuous signals over $\R^2$ and not discretized on~a~grid.}
 $f: \R^2 \to \R^c$ associating a $c$-dimensional \emph{feature vector} to each point on the plane $\R^2$.
The spatial structure of the feature maps implicitly defines the action of the translation group on them (by translating the points on the plane) and the use of convolution guarantees translation equivariance.
\begin{wrapfigure}{r}{0.5\textwidth}
	\centering
	\includegraphics[width=\linewidth]{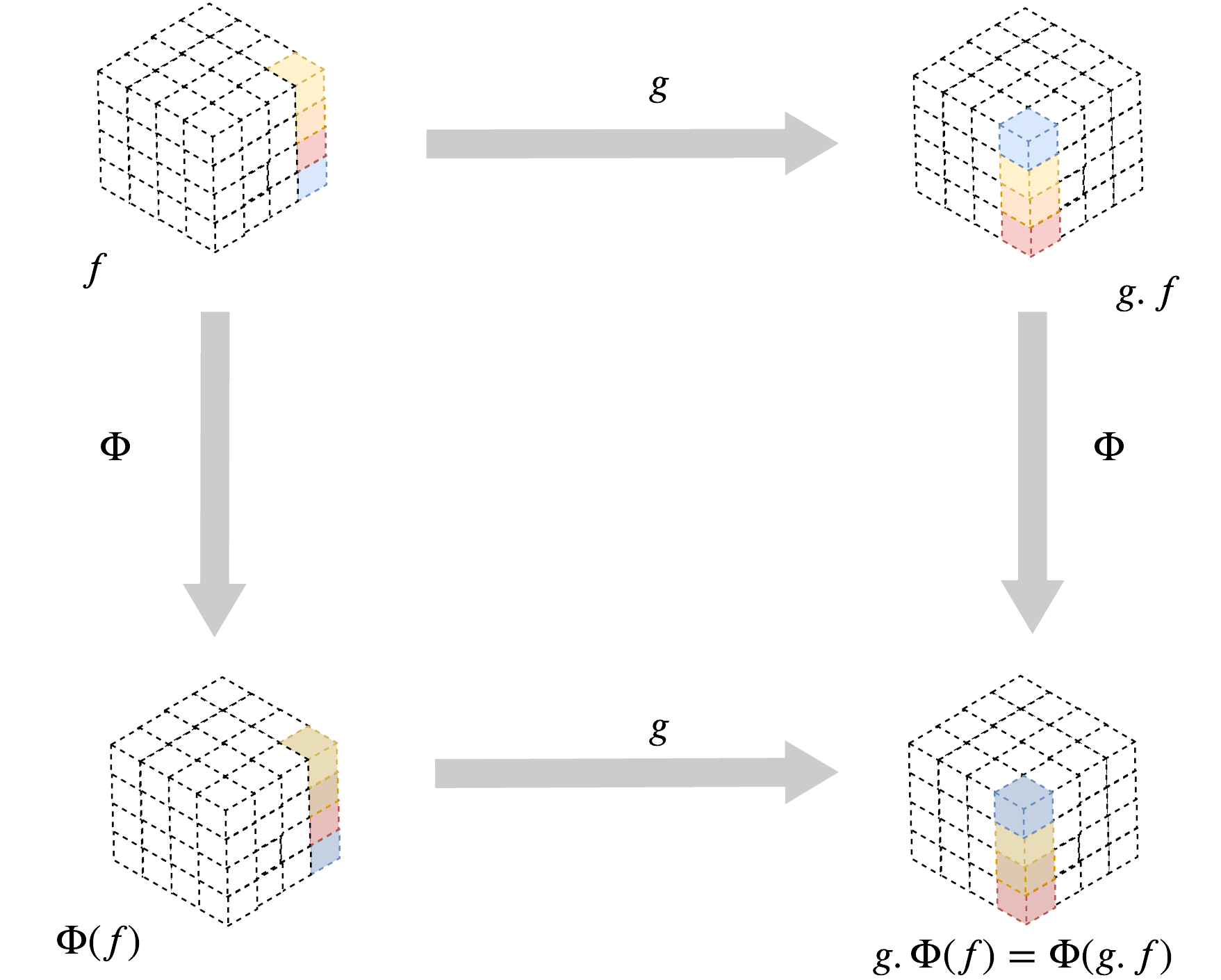}
	\caption{Action of the group of $4$ rotations on a convolutional feature $f$ (1st row): $g={\pi\over 2}$ moves the pixels on the plane and transforms each feature vector $f(x) \in \R^4$ by circularly shifting its channels. A similar action is defined on the output of a neural network $\Phi$ (2nd row). If $\Phi$ is equivariant, it commutes with these two actions (bottom right).}
	\label{fig:equivariance}	
	\vspace*{-2em}
\end{wrapfigure}
We are generally interested in exploiting a larger class of symmetries; for example, we can additionaly consider the group of $N$ discrete rotations containing $\{r{2\pi\over N}\}_{r=0}^{N-1}$.
Assuming translation equivariance is already guaranteed by the use of convolution, we denote with $G$ only the rotations.
More precisely, the $c$ channels of the feature field are split into $c/N$ blocks of size $N$. 
The group element $r{2\pi\over N}$, then, cyclically shifts the channels in the same block by $r$ positions.
The resulting equivariant model is effectively a group convolutional network.
Fig~\ref{fig:equivariance} shows the action of the group of $4$ rotations over a convolutional feature field with $c=N=4$ channels.

Enforcing rotation equivariance in a convolutional layer requires its filters to live in a lower dimensional vector space.
An equivariant filter can therefore be built by finding a basis for this space and learning the coefficients to linearly combine it.
For further details on the constraints on the layers of a network required by equivariance to the isometries of the plane, see~\cite{Weiler2019}. 
All the equivariant architectures are implemented using the \emph{PyTorch} library \emph{e2cnn}\footnote{\url{https://github.com/QUVA-Lab/e2cnn}}.

Unfortunately, because images are sampled on a pixel grid, only rotations by multiples of ${\pi\over2}$ are perfect symmetries.
In order to consider rotations by arbitrary angles smaller than ${\pi\over2}$, a common approach is defining filters in terms of a finite basis of steerable continuous filters \cite{Worrall2017-HNET,Weiler2018-STEERABLE,cheng2018rotdcf,Weiler2019}.
However, it is important that the continuous basis is band-limited: as discussed in \cite{Weiler2019}, sampling high-frequency filters on a grid can introduce non-equivariant elements in the basis due to aliasing.

%% file: sections/04-method.tex
%!TEX root=../iclr2020_conference.tex

\section{Method}
\label{sec:method}
\begin{wrapfigure}{l}{0.55\textwidth}
    \vspace*{-1.4em}
    \centering
    \includegraphics[width=\linewidth]{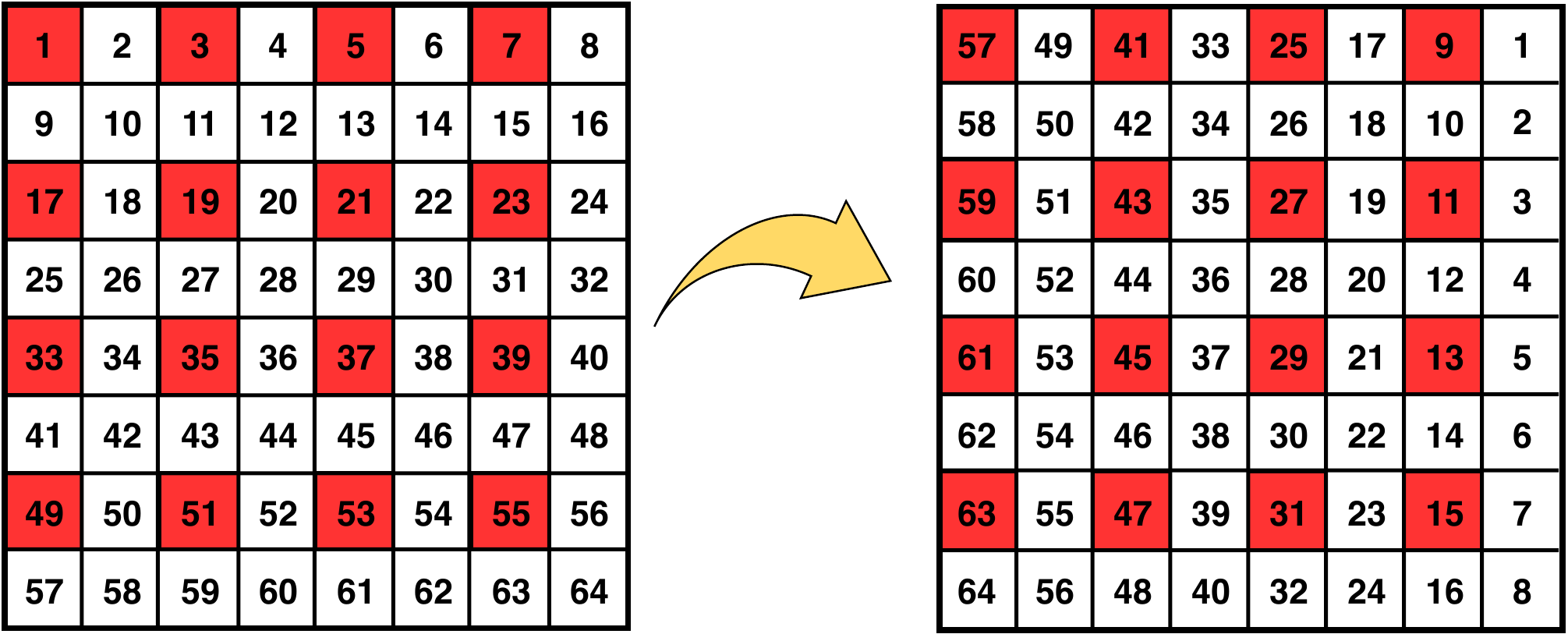}
    \caption{Strided convolution on an even-sized grid samples different points when the input is rotated, breaking equivariance.}
    \label{fig:strided_conv}
    \vspace*{-1.5em}
\end{wrapfigure}
In order to study how equivariance is affected when limited computational resources are available, we experiment with MobileNetV2~\cite{Sandler2018}.
To implement an equivariant version of MobileNetV2, we build an equivariant depth-wise convolution.
In our equivariant model, we preserve the same architecture, i.e. we use the same number of channels and the same filter sizes.
Rotations by angles $\theta < {\pi\over2}$ move the pixels in the corners outside the borders of the pixel grid.
In order to enforce the rotational symmetry of the images, we apply a circular binary mask to the input of a model.
Moreover, the global spatial pooling at the end of MobileNetV2 is combined with a similar mask to have a circular field of view.
The use of strided convolution can harm the stability of the model under rotations of the same input image.
For instance, Fig~\ref{fig:strided_conv} shows that a strided $3\times 3$ convolution over an even-sized feature map samples different points when the feature map is rotated by ${\pi\over2}$.
This was also noted in \cite{romero2020attentive}.
In order to preserve perfect ${\pi\over2}$ rotation equivariance, we use odd-sized inputs and adapt stride and padding of convolution layers to maintain odd-sized~pixel~grids.

\input{sections/04.2-quantization}

%% file: sections/04.2-quantization.tex
%!TEX root=../iclr2020_conference.tex

\subsection{Data-Free Quantization}
\label{subsec:quantization}

An efficient architecture design is often not enough to run deep networks on edge devices.
This means that other methods to reduce the computational costs are needed.
In this work, we use the \emph{data free quantization} (DFQ) methods from ~\cite{Nagel2019} which we show being compatible with an equivariant design.
Because an equivariant network can be converted to a conventional one after training, we can easily apply these techniques on our models. 
In particular, we will use \emph{cross-layer range equalization} and \emph{high bias absorption}.

\paragraph{Cross-layer range equalization}

Exploiting the commutative property of the ReLU function with respect to scaling of its input, it is possible to adapt the weights of a trained network for better quantization performance.
\cite{Nagel2019} use this property to equalize the range of values of the weights attached to each channel in a layer in order to maximize the channel precision after quantization.

Using the same notation as \cite{Nagel2019}, consider a pair of fully connected layers such that:
$$\yvec = W^{(2)} f(W^{(1)} \xvec + \bvec^{(1)}) + \bvec^{(2)} \ \ . $$
Equalization involves scaling the weights with a positive diagonal matrix $S$ as
\begin{equation}
\label{eq:equalization}
\yvec = W^{(2)} S f(S^{-1} W^{(1)} \xvec + S^{-1}\bvec^{(1)}) + \bvec^{(2)} 
\end{equation}
The diagonal of $S$ contains elements $s_i = \frac{\sqrt{r_i^{(1)} r_i^{(2)}}}{r_i^{(2)}}$, where $r_i^{(1)}$ and $r_i^{(2)}$ are respectively the ranges of values of the incoming weights $W^{(1)}_{i,:}$ and the outgoing weights $W^{(2)}_{:,i}$.
In a convolutional network, this is applied for each pixel and the matrices $W^{(*)}$ correspond to the convolutional kernels.
Assuming that the model is equivariant to $N=4$ discrete rotations and that the transformation law described in Sec~\ref{sec:equivariance} is used in all feature maps, this scaling is equivariant only if all the $N$ channels in the same block are scaled by the same factor.
Because of equivariance, the filters which map to different channels in the same block are rotations of each other and, therefore, share the same values.
A similar argument holds for the outgoing filters.
As a result, both terms $r_i^{(1)}$ and $r_i^{(2)}$ are shared within each $N$-dimensional block.
For $N>4$, this does not theoretically hold anymore because the rotation of the filters involves interpolation.
Nevertheless, because we apply band-limiting, rotated filters usually contain similar values.
Indeed, we observe that equivariance is only marginally affected in practice.

\paragraph{High-bias absorption}
The equalization performed in Eq.~\eqref{eq:equalization} scales also the bias.
This can increase the range of values it takes and, therefore, potentially, the ranges of the activations.
If the distribution of the inputs is concentrated in the positive domain of ReLU, it is possible to exploit its linear behavior to shift the input distribution and absorb part of the bias in the following layer.
Using batch normalization statistics, we can estimate the distribution and correct the bias without~additional~data.
Because an equivariant batch normalization layer shares statistics across each channel in the same block, it is guaranteed that the bias values of the channels in the same field are shifted by the same amount, preserving the equivariance.

%% file: sections/05-experiments.tex
%!TEX root=../iclr2020_conference.tex

\section{Experiments}
\label{experiments}

\begin{table}[th]
\caption{Test accuracy on PCam}
\label{tab:test_acc}
\begin{center}
\begin{tabular}{l|cc|c}
Model                    & Full-Precision   & Quantized (INT8)  & \\
\hline 
Conventional MobileNetV2 & $84.67 \pm 1.91$ &  $84.32 \pm 1.76$ & -0.4\% \\
Equivariant  MobileNetV2 & $89.19 \pm 0.79$ &  $88.94 \pm 0.66$ & -0.3\% \\
\hline 
Equivariant DenseNet \cite{Veeling2018-qh}  & 89.8              & - &- \\
\end{tabular}
\end{center}
\end{table}

We evaluate our classification models on the PatchCamelyon (PCam) dataset~\cite{Veeling2018-qh}.
It contains $96 \times 96$px images extracted from histopathologic scans and labeled based on the presence of metastatic tissue in their central $32 \times 32$px region.
The surrounding pixels do not influence the label but only provide context.
We consider two models: a conventional and an equivariant MobileNetV2, both adapted as described in the previous section for a fair comparison.
The input images are cropped to $95\times 95$px to have odd-size.
Moreover, we reduce the stride in two layers to adapt the model to the lower resolution of this dataset.
In the equivariant model, we consider the group of $12$ rotations as \cite{Weiler2018-STEERABLE,bekkers2018roto,Weiler2019} found only smaller improvements by using more.

We train both models for $300$ epochs and select the set of weights with the lowest loss on the validation set.
As in~\cite{Veeling2018-qh}, we augment the training set with ${\pi\over2}$ rotations and reflections.
To study the effect of quantization on equivariant models, we first apply the equalization techniques described in Sec~\ref{subsec:quantization} and then reduce the weights and activations precision to INT8.
The test accuracies of both the conventional and the equivariant models, before and after quantization are shown in Tab~\ref{tab:test_acc}.
As expected, enforcing equivariance leads to a significant improvements in the full-precision model.
The same improvement is preserved when the models are quantized, proving the quantization techniques applied are compatible with equivariance.

%% file: sections/06-conclusions.tex
%!TEX root=../iclr2020_conference.tex

\section{Conclusion}
Deep neural networks are known to require large amount of data to train and, often, complex and expensive architectures to obtain state-of-the-art performances.
However, the limited resources available in developing countries make deploying deep learning solutions challenging.
In this work, we try to solve this problem by combining two so far independent lines of research, quantization and equivariance, to achieve improved generalization and efficient inference.
In particular, we show that equivariant networks can be efficiently implemented and quantized without losing their desirable properties or reducing their expressive power.